\documentclass[10pt]{article}
\usepackage{fullpage}
\usepackage[utf8]{inputenc}

\title{\textbf{Analysis of draft EU ADS Performance Requirements}}
\author{Maria Soledad Elli\thanks{Intel Corporation {\tt\scriptsize \{maria.elli, jack.weast\}@intel.com}}\and Jack Weast\footnotemark[1]}

\date{\today}

\usepackage{graphicx}
\usepackage{xcolor}
\usepackage[hidelinks]{hyperref}
\usepackage{amsmath,amssymb}
\usepackage{xspace}
\usepackage{float}
\usepackage{caption}
\usepackage{subcaption}
\usepackage{wrapfig}
\usepackage{multicol}
\newcommand{\comment}[1]{}

\newcommand{\RSS}{\emph{RSS}\xspace}

\newcommand{\maxBrakeLong}{\beta^{\mathrm{lon}}_{\mathrm{max}}\xspace}
\newcommand{\maxAccelLong}{\alpha^{\mathrm{lon}}_{\mathrm{max}}\xspace}
\newcommand{\minBrakeLong}{\beta^{\mathrm{lon}}_{\mathrm{min}}\xspace}

\newcommand{\maxAccelLat}{\alpha^{\mathrm{lat}}_{\mathrm{max}}\xspace}
\newcommand{\minBrakeLat}{\beta^{\mathrm{lat}}_{\mathrm{min}}\xspace}

\begin{document}

\maketitle

\section{Introduction}

Recently, the European Commission published draft regulation for uniform procedures and technical specification for the type-approval of motor vehicles with an automated driving system (ADS) \cite{EUdraftreg}, \cite{EUdraftreg_annex}.  While the draft regulation is welcome progress for an industry ready to deploy life saving automated vehicle technology, we believe that the requirements can be further improved to enhance the safety and societal acceptance of automated vehicles (AVs).
In this paper, we evaluate the draft regulation's performance requirements that would impact the Dynamic Driving Task (DDT). We highlight potential problems that can arise from the current proposed requirements and propose practical recommendations to improve the regulation.



\section{Draft EU Regulation Performance Requirements for the DDT}
In this section, an analysis of the performance requirements on an ADS contained in the draft regulation is presented. Where potential issues with the draft regulation have been found, proposed solutions are presented.

\subsection{Required vehicle behaviour under normal driving conditions}
Annex II, Section 3 of the draft regulation annex \cite{EUdraftreg_annex} lists a number of  performance requirements for the expected vehicle behaviour regarding the DDT under normal driving conditions. A subset of those requirements relevant for the analysis presented in this paper are re-printed here:

\begin{itemize}
\item "In the ODD, the ADS shall perform the DDT safely and dynamically by appropriate choice of trajectory and speed and shall manage all normal driving situations reasonably expected in the ODD."
\item "The ADS shall be able to leave sufficient space with the vehicle in front to avoid a collision. In case this cannot be respected temporarily because of other road users (e.g. vehicle is cutting in, decelerating lead vehicle, etc.), the vehicle shall readjust the following distance at the next available opportunity"
\item "When travelling on its lane the ADS shall be able to leave sufficient time and space for others in  lateral  manoeuvres  as  appropriate."
\item "The ADS shall be cautious with right of way at intersections"
\item "The ADS shall drive in a predictable manner for other road users and shall avoid any harsh braking/steering unless an emergency manoeuvre would become necessary."
\end{itemize}

The draft regulation does not define what constitutes "sufficient space with the vehicle in front to avoid a collision", "sufficient time and space in lateral manoeuvres", "be cautious with right of way", and/or "harsh braking/steering", leaving such requirements open to wildly differing interpretations by different ADS developers. Fortunately, \textcolor{black}{previous work, such as \cite{shalev2017formal}}, can be used to formally define such concepts so that industry can be given clarity on expectations for performance requirements.

First, the requirement to maintain a "sufficient space with the vehicle in front to avoid a collision" can be formally defined following \textcolor{black}{the minimum safe distance definition from \cite{shalev2017formal}}, as shown in equation \eqref{eq:rss_same_dir}.

\begin{equation}\label{eq:rss_same_dir}
       d^{\mathrm{lon}}_{\mathrm{min}} = \Bigg[ v_r\rho + \frac{1}{2}\maxAccelLong\rho^2 + 
       \frac{ (v_r+ \maxAccelLong \rho)^2}{2\minBrakeLong} - \frac{v_f^2}{2\maxBrakeLong} \Bigg]_+
\end{equation}

Where $\mathrm{[x]_+  := max\{x,0\}}$, 

$v_r$ and $v_f$ are the rear vehicle and the lead vehicle's longitudinal velocity respectively.

$\rho$ is the response time of the rear vehicle.

$\maxAccelLong$ is the maximum acceleration that the rear vehicle could apply during its response time $\rho$.

$\minBrakeLong$ is the minimum deceleration the rear vehicle can apply, with $\minBrakeLong \le \maxBrakeLong$.

$\maxBrakeLong$ is the maximum assumed reasonably foreseeable deceleration the front vehicle can apply.

\begin{figure}[H]
\centering
\includegraphics[width=0.7\textwidth]{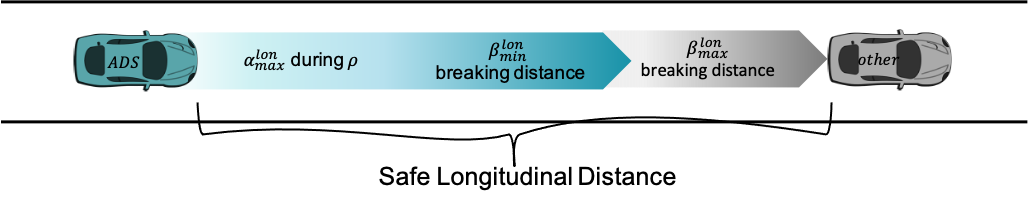}
\caption{ADS Safe Longitudinal Distance}
\label{fig:rss_safe_d_lon}
\end{figure}

This formally verified, physics-based minimum safe distance definition accounts for reasonable worst-case braking of the front vehicle as well as other important aspects such as the time it may take the ADS to respond to the situation (see figure \ref{fig:rss_safe_d_lon}). 
If the minimum safe distance defined by \eqref{eq:rss_same_dir} is compromised, then the ADS should apply a deceleration of at least $\minBrakeLong$ in order to avoid a collision. \\

\begin{figure}[H]
\centering
\includegraphics[width=0.3\textwidth]{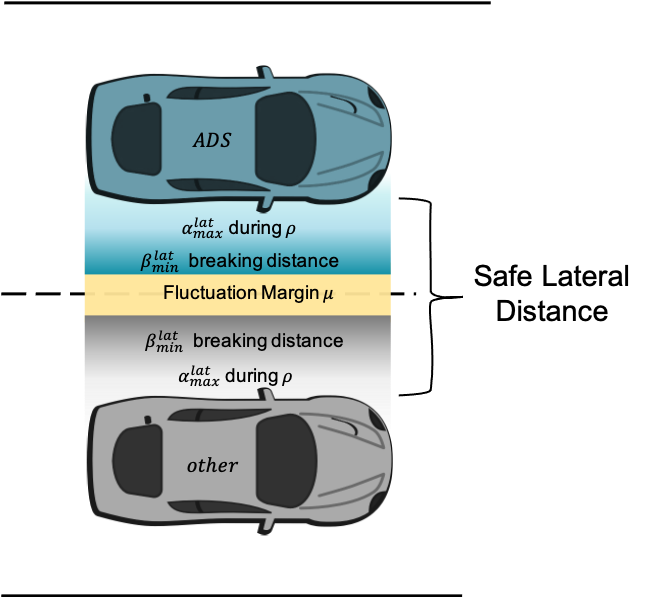}
\caption{ADS Safe Lateral Distance}
\label{fig:rss_safe_d_lat}
\end{figure}

Second, the regulations requirement to have "sufficient time and space in lateral manoeuvres" can be defined following the \emph{minimum safe lateral distance} definition from \cite{shalev2017formal}. 
In order for the ADS to make a lateral manoeuvre, it needs to have enough lateral space (as defined by equation \eqref{eq:rss_lat}) with another vehicle such that it can avoid a collision if both vehicles were to steer towards each other during their response time and then steer away (see figure \ref{fig:rss_safe_d_lat}).\\

\begin{equation}\label{eq:rss_lat}
d^{\mathrm{lat}}_{\mathrm{min}}  = \mu + \left[\frac{v_1 + v_{1,\rho}}{2}\rho_1 +
\frac{v_{1,\rho}^2}{ 2 \minBrakeLat} - \left( \frac{v_2 + v_{2,\rho}}{2}\rho_2 - \frac{v_{2,\rho}^2}{ 2
   \minBrakeLat} \right) \right]_+
\end{equation}

with $v_{1,\rho_1} = v_1 + \rho_1 \maxAccelLat$ and $v_{2,\rho_2} = v_2 + \rho_2 \maxAccelLat$,

$v_1$ and $v_2$, are the lateral velocities of vehicle 1 to the left of vehicle 2.
 
 $\rho_{1,2}$ are vehicle's response time.
 
$\maxAccelLat$ is the maximum lateral acceleration towards each other that vehicles could have during their response time.

$\minBrakeLat$ the minimum lateral deceleration that vehicles can apply to move away from each other until they reach zero lateral velocity.

$\mu$ is a safe lateral margin that accounts for small lateral fluctuations of the vehicles.

\begin{figure}[H]
\centering
\includegraphics[width=0.6\textwidth]{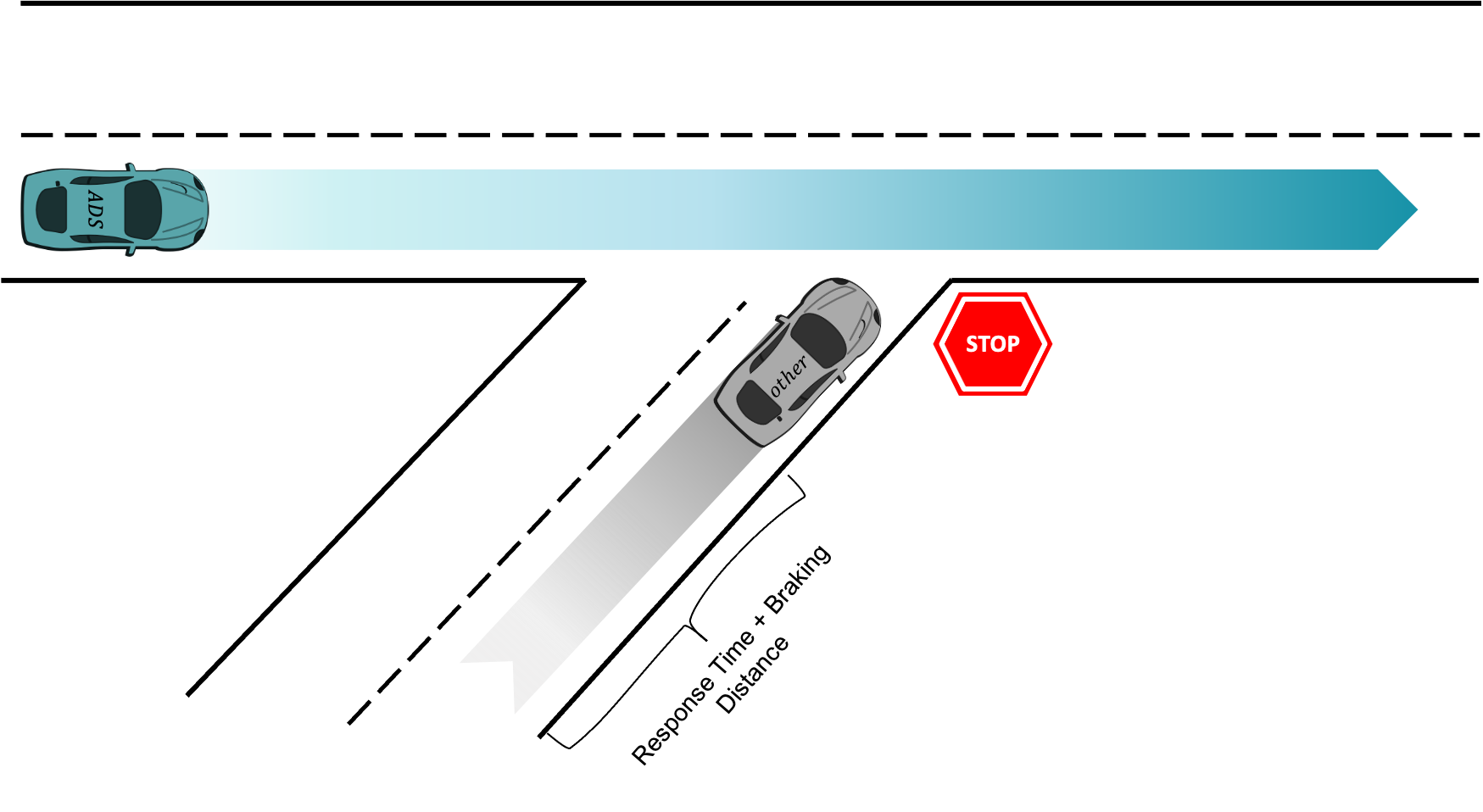}
\caption{Safe Stopping Distance of non-prioritized road users}
\label{fig:rss_intersection}
\end{figure}

Third, definitions of being "cautious with right of way at intersections" can follow the common sense rule "right of way is given, not taken". 
Human drivers usually prefer to be safe rather than correct, and so even though a prioritized vehicle may have the right of way, it may smartly yield to another vehicle in order to avoid a collision. 
In order to define cautiousness at intersections, a simple notion of when to give way can be defined as "a vehicle which has to give way should be able to stop safely before encroaching the prioritized lane" \cite{shashua2018implementing}. 
This means that the non prioritized vehicle has to maintain a safe distance with respect to the intersection point of both vehicles' routes, such that after stopping, it will not encroach upon the prioritized vehicle's route (see figure \ref{fig:rss_intersection}). 

The distance that a non-prioritized vehicle that is traveling with velocity $v$ has to maintain in order not to violate the right of way can be defined as in equation \eqref{eq:rss_braking}

\begin{equation}\label{eq:rss_braking}
       d^{\mathrm{lon}}_{\mathrm{brake}} = \Bigg[ v\rho + \frac{1}{2}\maxAccelLong\rho^2 + 
       \frac{ (v + \maxAccelLong \rho)^2}{2\minBrakeLong} \Bigg]_+
\end{equation}

Where:

$\rho$ is the response time of the vehicle.

$\maxAccelLong$ is the maximum acceleration that the vehicle could apply during its response time $\rho$.

$\minBrakeLong$ is the minimum deceleration the vehicle can apply in order to respect the right of way.

Fourth, the failure to define "harsh braking/steering" could also be problematic, with different manufacturers making different assumptions. 
Taking a theoretical worst case assumption (e.g., the possible presence of a race car with 4g deceleration) will result in overly conservative worst case behavior, failing to conform with any definition of what is reasonably foreseeable. 
Clearly defining such boundaries to know what is reasonably foreseeable vs. what is unexpected and thus may require an emergency evasive manoeuvre is needed in order for ADS-equipped vehicles to be able to drive in a predictable manner that conforms with societal expectations. 
Once such boundaries are defined, they can be used as inputs for the parameters $\maxBrakeLong$ and $\minBrakeLat$ of equations [\eqref{eq:rss_same_dir}, \eqref{eq:rss_braking}] and \eqref{eq:rss_lat} respectively. 


\subsection{Specific requirements regarding the Lane Change Procedure (LCP)}
\label{sec:req_lcp}
In the case of a Lane Change Procedure (LCP), the draft regulation states that the ADS may undertake a LCP only if:
\begin{itemize}
\item "The vehicle with the ADS would be able to keep a safe distance from a lead vehicle or any other obstacle in the target lane and if an approaching vehicle in the target lane is not forced to unmanageably decelerate due to the lane change of the vehicle with automated driving function."
\item "An approaching vehicle in the target lane should always have a TTC to the vehicle with automated driving function of at least 4 seconds at the end of the Lane Change Manoeuvre (LCM)"
\item "At the beginning of the LCM, the distance between the rear of the vehicle with automated driving function and the front of a vehicle following behind in the target lane at equal or lower longitudinal speed shall never be less than the speed which the following vehicle in target lane travels in 1 second"
\end{itemize}

\begin{figure}[H]
\centering
\includegraphics[width=0.7\textwidth]{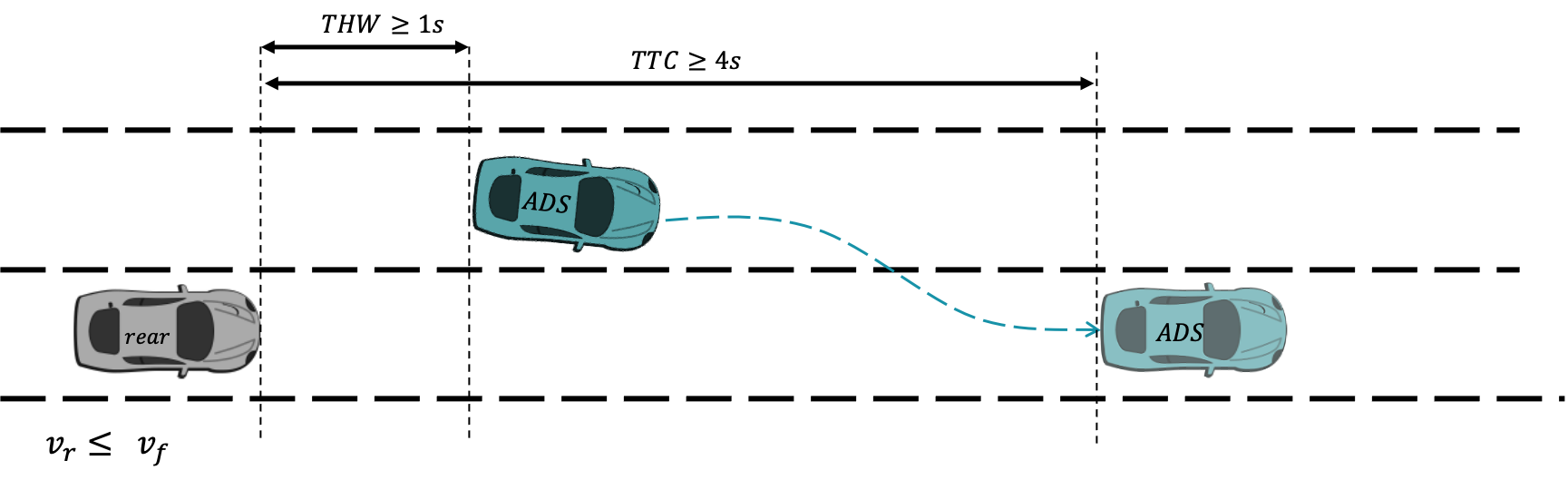}
\caption{Diagram of Lane Change Procedure Requirements}
\label{fig:lcp_diagram}
\end{figure}

A diagram of the draft regulation's requirements for a LCP  initiated by the ADS are depicted in figure \ref{fig:lcp_diagram}, under the assumption (though not explicitly defined in the regulation) that the "beginning of the Lane Change Manoeuvre" is when the ADS-equipped vehicle begins its lateral motion towards the target lane and "end of the Lane Change Manoeuvre" is the moment when the ADS has reached a lateral speed of $\approx 0 m/s$ in the target lane. 

The time to collision (TTC) \cite{hayward1972near} requirement of 4 seconds at the end of the lane change manoeuvre can be defined as in equation \eqref{eq:ttc}, with $distance$ being the headway distance between the approaching vehicle (i.e., rear vehicle) and the ADS vehicle (i.e., front vehicle), traveling with speeds $v_r$ and $v_f$, respectively. 

\begin{equation}
    TTC = \frac{distance}{v_r - v_f}\ge 4s
    \label{eq:ttc}
\end{equation} 

The time headway (THW) requirement for initiating a LCP with respect to a vehicle that may already be occupying the target lane, can be defined in  \eqref{eq:thw}.
\begin{equation}
    THW = \frac{distance}{v_r}\ge 1s
    \label{eq:thw}
\end{equation} 

It is important to highlight that an arbitrary and fixed TTC threshold does not scale well with different vehicle speeds as the formulation is \textit{only} dependent on the difference in the vehicles' speed (i.e., $\Delta_{speed} = v_r - v_f$).  In other words, a TTC of 4 seconds may make sense at high speeds where vehicles can cover a large amount of roadway in a manner of seconds and considerable time and distance may be required to come to a complete stop; however, when applied to vehicles operating at extremely low speeds where the vehicle could come to a complete stop in less than 1 second, preserving a TTC of 4 seconds would result in overly conservative behavior.  A fixed TTC value also does not consider the impact that road conditions (e.g. wet, snow) can have on the braking capability of a vehicles.

In order to analyze the minimum required distance be at the end of the LCP, we derive the expression defined by equation \eqref{eq:ttc_distance_req}.

\begin{equation}
    distance_{TTC} \ge 4*(v_r-v_f)
    \label{eq:ttc_distance_req}
\end{equation} 

This expression defines that, for a difference on vehicles’ speeds of $\Delta_{speed} = 1 m/s$, the required minimum distance would be 4 meters. It is important to note that this minimum TTC-based distance requirement is the same for vehicles traveling with $v_r = 5 m/s$ and $v_f=4 m/s$ as well as for vehicles traveling with $v_r = 25 m/s$ and $v_f=24 m/s$. 

Therefore, if both vehicles are driving at relatively low speeds, a 4 meters distance may be enough in order to avoid a collision if the leading vehicle were to suddenly brake. 
But, as we will see, if the vehicles are driving at higher speeds, a 4 meters of a headway distance may not be enough to avoid an accident if the front vehicle were to suddenly reduce its speed. 


More problematic, in the case that both vehicles are driving at the same speed, the safe distance requirement according to \eqref{eq:ttc_distance_req} is zero due to the $\Delta_{speed}$ being zero - regardless of how physically close the vehicles are.
While a default gap can be added in cases where $\Delta_{speed} = 0$, a fixed distance gap may not be sufficient when vehicles are driving at high speeds, or the opposite, excessive when vehicles are driving at low speeds. 

\begin{figure}[H]
\centering
\includegraphics[width=0.6\textwidth]{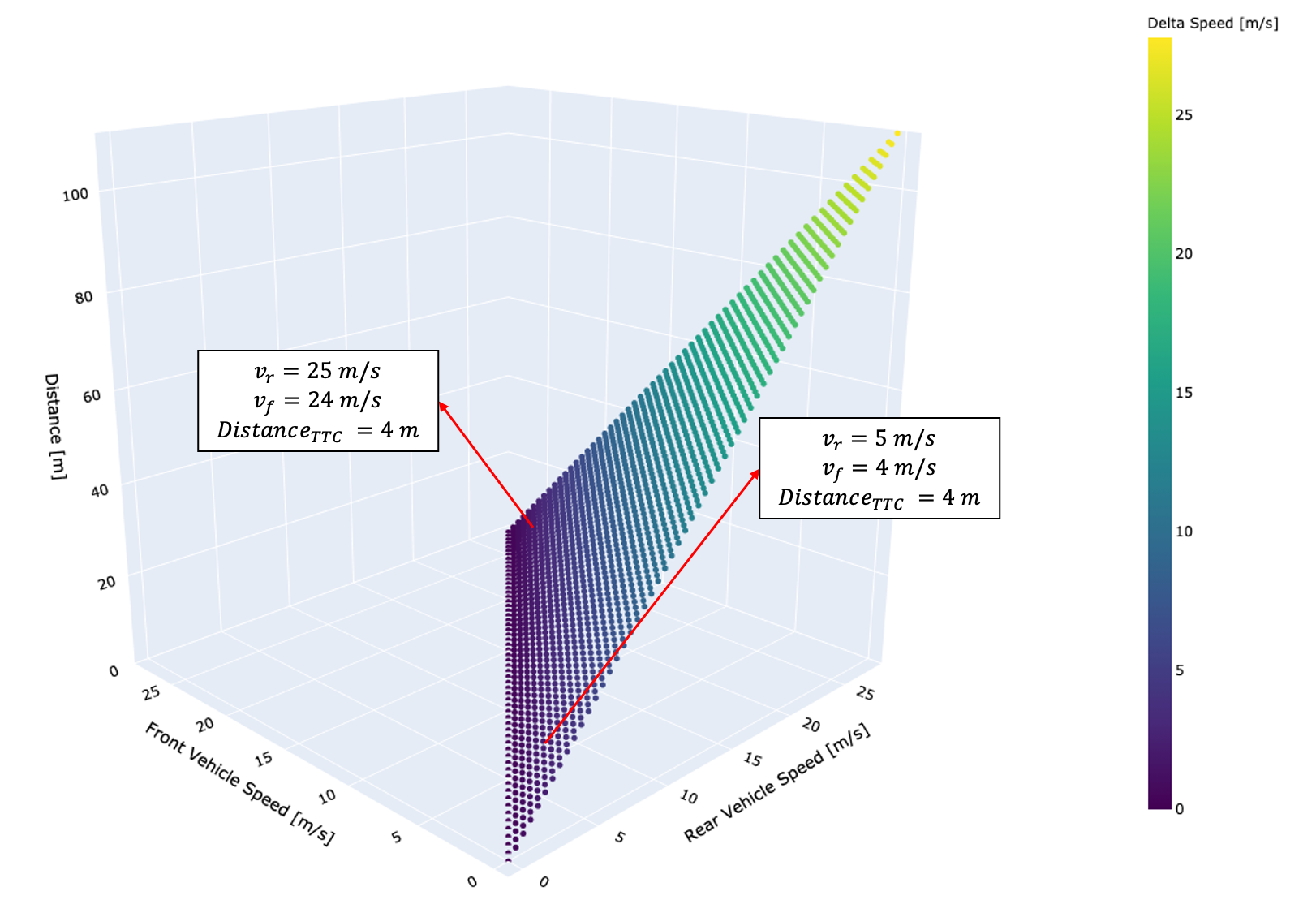}
\caption{Required Distance at end of LCP ($z$ axis) at different $v_r$ and $v_f$}
\label{fig:ttc_distance}
\end{figure}

Figure \ref{fig:ttc_distance} depicts the distance requirement from equation \eqref{eq:ttc_distance_req} assuming vehicles' maximum speeds of 100 km/h (27.7 m/s). In the figure we can see that the distance requirement increases ($z$ axis) as the $\Delta_{speed}$ increases (color bar). 

Table \ref{tab:ttc_distance} shows the distance that an ADS-equipped vehicle would have to achieve at the end of a LCP with an approaching vehicle in the target lane that is driving with $v_r =  50\ km/h$ (a typical urban speed limit). The equivalent required car lengths (assuming a car length of 4.7m) is shown in the table for comparison.

\begin{table}[H]
\begin{center}
 \begin{tabular}{||c c c c c c||} 
 \hline
 $v_r$  & $v_f$  & $\Delta_{speed}$  & $\Delta_{speed}$  &	$distance_{TTC}$ & Car lengths \cr
 [km/h] &  [km/h] &  [km/h] &  [m/s] & [m] & \\
 \hline\hline
\textcolor{black}{50} & \textcolor{black}{50} & \textcolor{black}{0} & \textcolor{black}{0} & \textcolor{black}{0} & \textcolor{black}{0}\\
\hline
50 & 40 & 10 & 2.8 & 11.1 & 2.3\\
\hline
50 & 30 & 20 & 5.6 & 22.2 & 4.7\\
\hline
\textbf{50} & \textbf{20} & \textbf{30} & \textbf{8.3} & \textbf{33.3} & \textbf{7.0}\\
\hline
50 & 10 & 40 & 11.1 & 44.4 & 9.4\\
\hline
\end{tabular}
\end{center}
\caption{Minimum distance required at $TTC \ge 4$ at different vehicles speeds}
\label{tab:ttc_distance}
\end{table}

As can be seen in the table, the proposed regulation would require potentially extreme car length gaps to be created when performing a lane change manoeuvre, rendering the performance of the ADS-equipped vehicle unpractical in the real world.

\begin{figure}[H]
\centering
\includegraphics[width=1\textwidth]{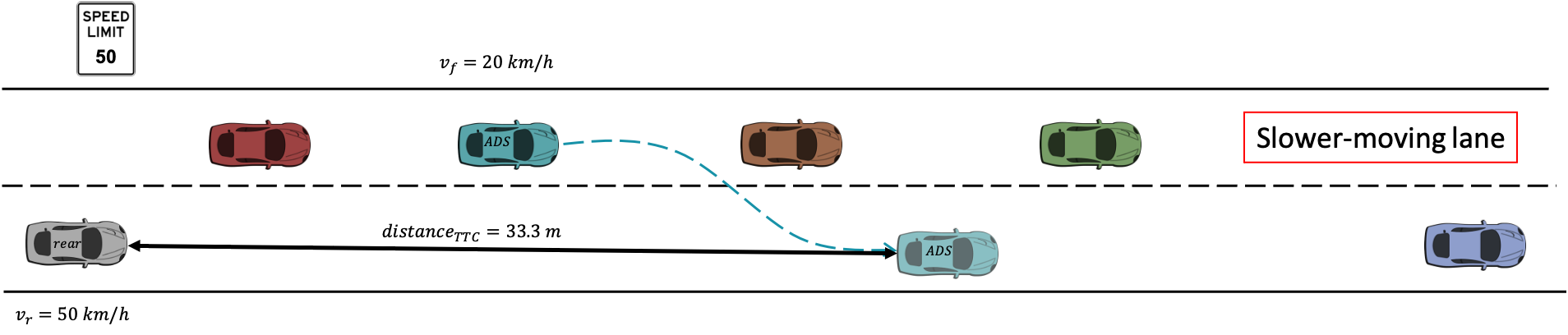}
\caption{Example of required distance for ego vehicle to perform LCM in urban scenario}
\label{fig:ttc_distance_example}
\end{figure}

Consider the urban driving scenario depicted in figure \ref{fig:ttc_distance_example}.
The ADS-equipped vehicle finds itself driving on a slower-moving lane and is planning on changing lanes. 
If the ADS is driving at 20 km/h, it would need to find a 33.3 m gap to perform the lane change manoeuvre in order to avoid interfering with the approaching vehicle that is driving at 50km/h. 
Considering that most city blocks usually have a length of $\approx$100m between intersections \cite{daley2007street}, requiring nearly a third of a city block's worth of distance for the ADS to perform a lane change would be unpractical at best.

\subsubsection{Proposed Enhancements to LCP Requirements}
As has been shown, a requirement to maintain a fixed TTC value does not scale well with different vehicles' speeds and can pose serious risks to the road safety as well as negatively impacting  traffic flow which will delay and slow the adoption of this life-saving technology. 
A time-based requirement, therefore, needs to be dynamic and must adapt to different speeds and consider the actual braking capabilities of the vehicles in order for it to be useful. 

Beginning with  equation \eqref{eq:rss_same_dir}, we can derive a dynamic TTC based threshold; where $distance$ is the minimum longitudinal safe distance, if $\rho = 0$: 
\begin{equation*}
    distance \ge \Bigg[ \frac{ v_r^2}{2\minBrakeLong} - \frac{v_f^2}{2\maxBrakeLong} \Bigg]_+
\end{equation*}
if $\maxBrakeLong = \minBrakeLong$:
\begin{equation*}
\begin{split}
     distance \ge \Bigg[ \frac{ v_r^2}{2\maxBrakeLong} - \frac{v_f^2}{2\maxBrakeLong} \Bigg]_+\\ 
     distance \ge \Bigg[ \frac{ v_r^2 - v_f^2}{2\maxBrakeLong} \Bigg]_+\\ 
     distance \ge \frac{ (v_r - v_f)(v_r + v_f)}{2\maxBrakeLong} \\ 
     \end{split}
\end{equation*}

\begin{equation}\label{eq:dynamic_ttc}
     TTC = \frac{distance}{(v_r - v_f)} \ge \frac{(v_r + v_f)}{2\maxBrakeLong} := dynamic_{TTC}
\end{equation}

In this way, a dynamic TTC threshold that considers the vehicles' braking capabilities and that depends on the average speed of the vehicles can be derived, making it a more accurate measure of safety than \eqref{eq:ttc}. 
Having such a dynamic threshold requirement can greatly improve traffic efficiency while at the same time provide clear safety margins for the ADS to maintain. 

\begin{figure}[H]
\centering
\includegraphics[width=0.6\textwidth]{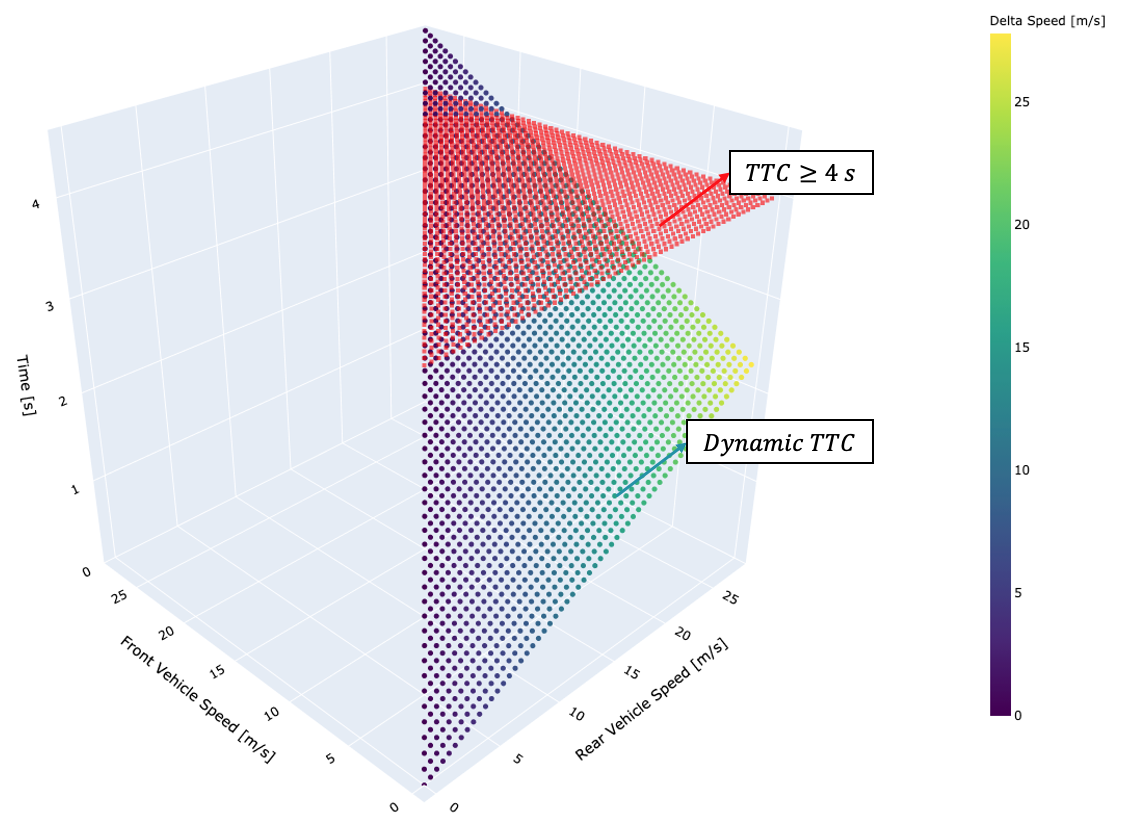}
\caption{Fixed TTC threshold of 4 s (red plane) compared to a dynamic TTC threshold at different $v_r$ and $v_f$, assuming $\maxBrakeLong = 6 m/s^2$}
\label{fig:dynamic_ttc}
\end{figure}

Figure \ref{fig:dynamic_ttc} shows the dynamic time-to-collision requirement for a vehicle performing a LCP changes based on the speeds of both vehicles. Assuming $\maxBrakeLong = 6 m/s^2$, when both vehicles are driving at 86 km/h (24 m/s), the minimum $dynamic_{TTC}$ requirement is 4s. 
However, when the vehicles are driving at a speed of 18 km/h (5 m/s), the $dynamic_{TTC}$ requirement is only 0.8s. 
Intuitively, this makes sense as the amount of time and distance necessary to come to a complete stop and avoid a collision is much smaller at lower speeds.
 
While equation \eqref{eq:dynamic_ttc} provides a dynamic method to calculate the TTC threshold, the response time of the ADS is neglected (or rather, assumed to be zero). Therefore, a better definition would be one that considers any delay in the ADS' response, as defined in equation \eqref{eq:dynamic_ttc_w_rho}.

\begin{equation}\label{eq:dynamic_ttc_w_rho}
     TTC \ge \frac{(v_r + v_f)}{2\maxBrakeLong} + \rho := dynamic_{TTC}
\end{equation}

This analysis underscores how important it is that performance requirements for ADS-equipped vehicles must be dynamic and take into consideration all relevant variables, including the braking capability and reaction time of the ADS-equipped vehicle, but also the reasonable worst-case behavior of others.\\

\subsection{Specific requirements regarding turning and crossings}
The draft regulation also contains performance requirements for intersections and specifically scenarios where the ADS-equipped vehicle may be crossing a lane used by vehicles traveling in opposite directions.

\begin{figure}[H]
\centering
\includegraphics[width=0.5\textwidth]{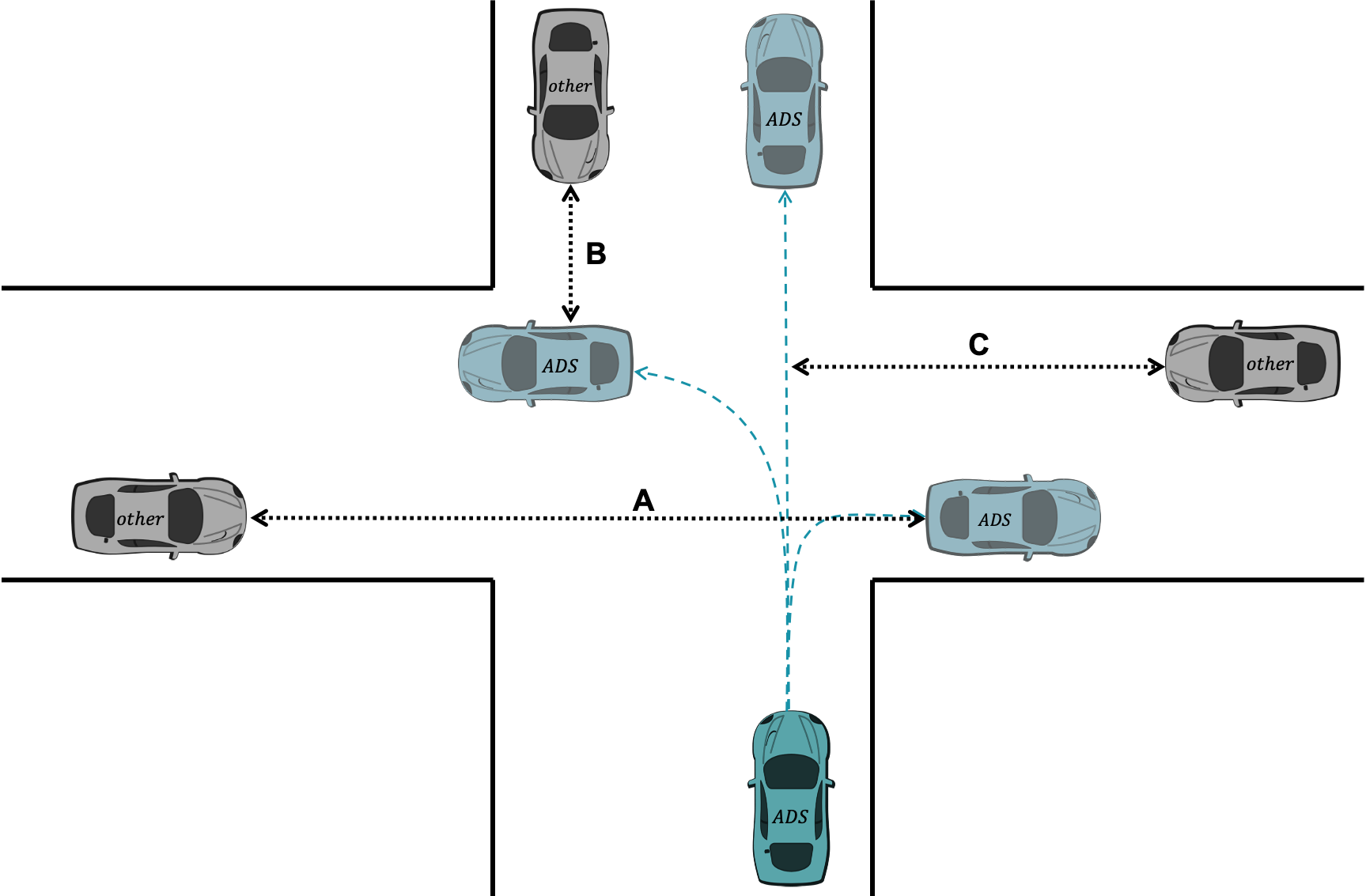}
\caption{Diagram of the distances during turning and crossings. Case (A): Distance to the following traffic to be observed during turning-in. Case (B): Additional distance to reverse traffic to be observed when turning as a result of reverse traffic. Case (C): distance to the privileged crossing traffic to be respected when crossing}
\label{fig:intersection}
\end{figure}

The regulation states:
\begin{itemize}
\item "In the case of the turning manoeuvre crosses the opposite track, when considering oncoming traffic, it must be ensured — in addition to the distance from the subsequent traffic on the target road — that the TTC of the privileged opposite traffic to the fictitious collision point (point of intersection of the trajectories) never falls below 4 seconds. (case (b) in Figure \ref{fig:intersection})"
\item "The same applies to cross with privileged traffic (case (c) in Figure \ref{fig:intersection}): The TTC of privileged traffic to the imaginary collision point (point of intersection of the trajectories) shall be greater than 4 seconds"
\end{itemize}

At intersection situations, case \textbf{A} from figure \ref{fig:intersection} is equivalent to the Lane Change Procedure and so would require the ADS-equipped vehicle to achieve a TTC of at least 4 seconds with respect to the vehicle following behind on the target road, meaning all the same problems highlighted in section \ref{sec:req_lcp} apply here as well. 
Depending on vehicles' speed, a 4 seconds threshold could be overly conservative or not enough in case the merging ADS-equipped vehicle suddenly reduces its speed due to slow traffic or other obstruction that may have been exposed ahead. 

For cases \textbf{B} and \textbf{C}, the distance requirement of the privileged vehicle (notated as the "other" vehicle in the figure) with respect to a fictitious collision point between the two paths is equivalent to $distance_{TTC} = 4*v_{other}$. 
Therefore, the distance required between the other vehicle and the fictitious point of intersection with the ADS-equipped vehicle's path should always be greater than 4 times the speed in $m/s$ of the other vehicle. 

Consider, for example case \textbf{B}, where the ADS-equipped vehicle is turning left at an intersection with traffic traveling towards the intersection at 30 km/h (8.3 $m/s$). 
The regulation's requirement means that the ADS-equipped vehicle cannot perform the left turn across the conflict point \textbf{B} unless there is a distance of 33.2 meters between the other vehicle and the fictitious intersection point with the ADS-equipped vehicle's path. 
The same would happen in case \textbf{C}, where the ADS-equipped vehicle is crossing the intersection with privileged traffic crossing. 

The performance requirement in the current draft regulation would unnecessarily restrict an ADS-equipped vehicle's ability to negotiate and perform a routine right or left turns in urban driving situations where the traffic density is high and such large distance gaps are uncommon.




\subsubsection{Proposed enhancements for requirements regarding turning and crossing}

Safety margins at intersections must allow for natural traffic flow and an enhancement over the current draft regulation can be realized using a common sense rule "Right of way is given, not taken". 
Following definitions from \cite{shalev2017formal} in cases of intersections, when a non-prioritized vehicle approaches an intersection with oncoming traffic, the ADS should maintain a longitudinal distance with respect to the conflict point such that it will not unsafely encroach upon the prioritized vehicle's path.

The specific time requirement can be realized by considering the response time $\rho$ of the prioritized vehicle travelling with speed $v$ and the time it would take it to stop at a given deceleration, $\minBrakeLong$ (see \eqref{eq:ttc_intersection}).

\begin{equation}\label{eq:ttc_intersection}
     TTC \ge \frac{v}{2\minBrakeLong} + \rho := TTC_{intersection}
\end{equation}



\begin{figure}[H]
\centering
\includegraphics[width=0.5\textwidth]{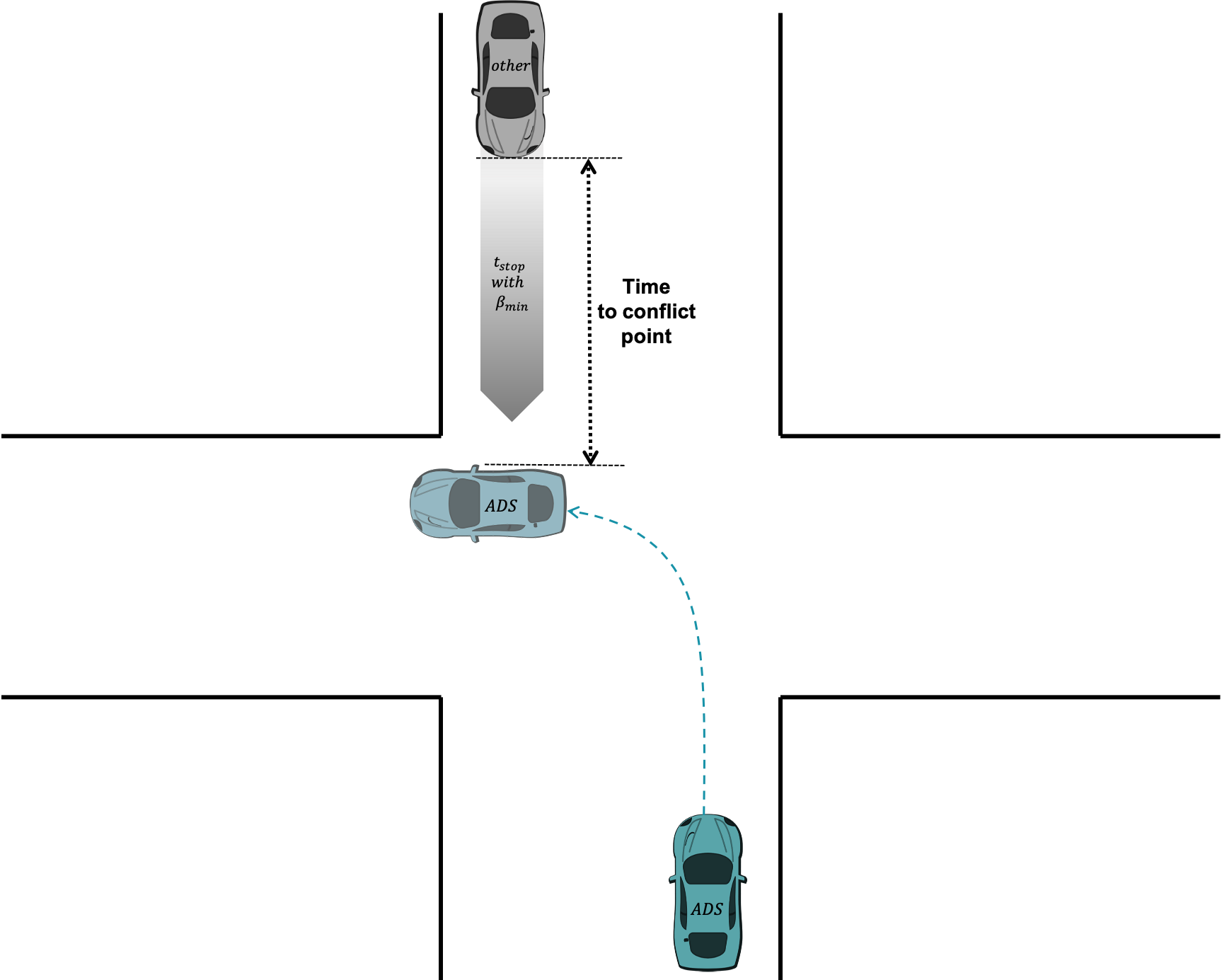}
\caption{ADS performing a safe unprotected left turn}
\label{fig:rss_ttc_intersection_b}
\end{figure}

In this way, the non-prioritized vehicle abiding by "Right of way is given, not taken" is able to navigate through the intersection and take the right of way only when the time of the prioritized vehicle to the conflict point is greater than that defined by \eqref{eq:ttc_intersection} (as shown in figure \ref{fig:rss_ttc_intersection_b}). In this way, the non-prioritized vehicle (i.e., ADS) is able to negotiate right of way through the intersection without forcing the prioritized vehicle into an unwanted braking manoeuvre.

\begin{figure}[H]
\centering
\includegraphics[width=0.45\textwidth]{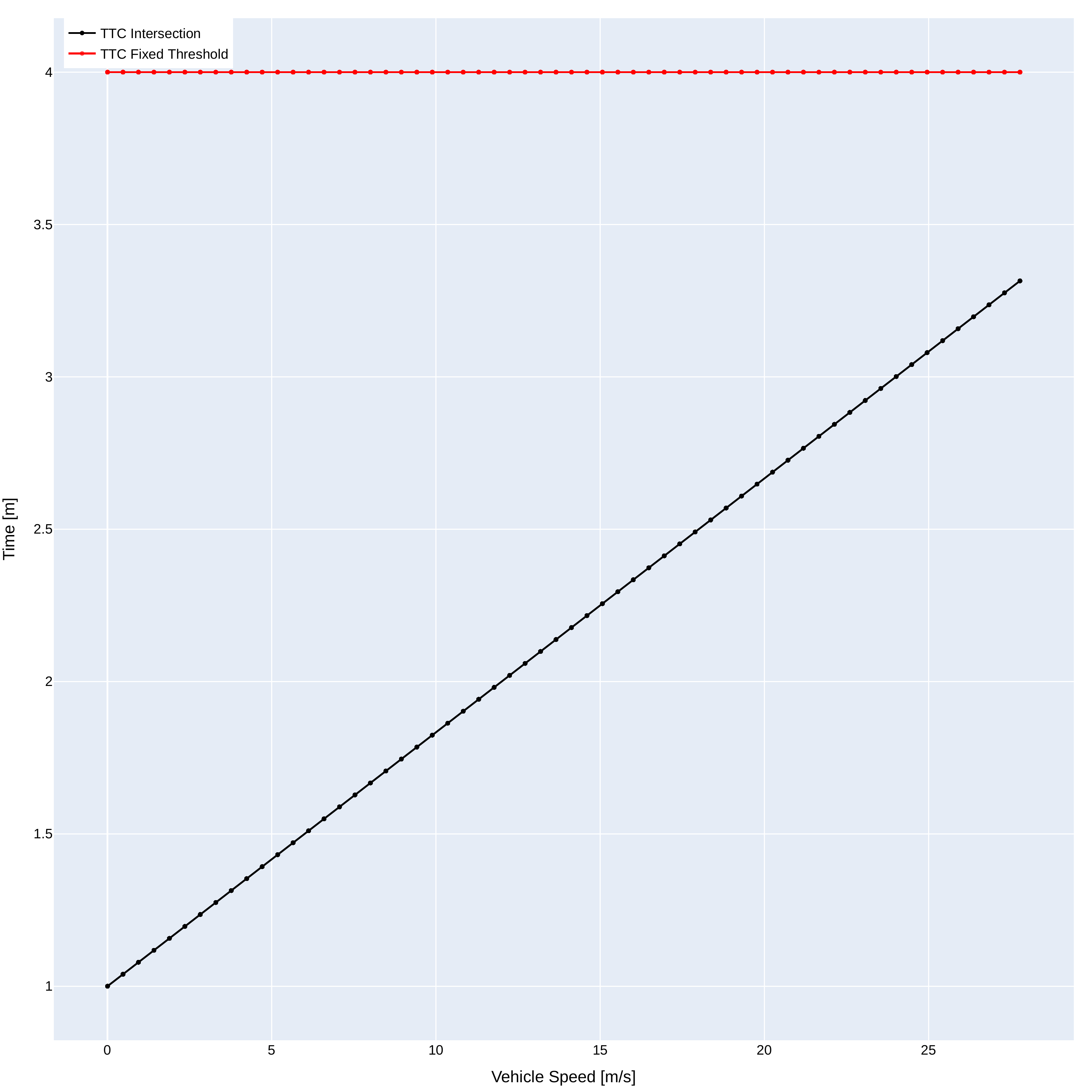}
\caption{Dynamic $TTC_{intersection}$ requirement compared to fixed TTC threshold requirements for ADS at intersections}
\label{fig:rss_ttc_intersection}
\end{figure}

This definition allows the ADS to navigate through the intersection assertively, but \textit{safely}. We can see the contrast on both approaches, namely a fixed TTC requirement and a dynamic $TTC_{intersection}$ requirement, in Figure \ref{fig:rss_ttc_intersection} that shows the time required by equation \eqref{eq:ttc_intersection}, (assuming $\minBrakeLong = 6 m/s^2$, and $\rho=1 s$), and the fixed TTC of 4 seconds. 
If the other vehicle is approaching the intersection at 30 km/h (8.3 m/s) the requirements from \eqref{eq:ttc_intersection} would not allow the ADS to go through the intersection unless the other vehicle has a TTC of at least 1.7s from the intersection point, compared to the 4s required by the current regulation requirements.

At nearly all speeds, the proposed enhancement provides for more optimal traffic flow while still supporting the same level of safety.

\subsection{Expected vehicle traffic behaviour in emergency conditions: cut-in}

Finally, the draft regulation requires the ADS to avoid collision with a cut-in vehicle if:
\begin{itemize}
    \item "Collisions with cutting in vehicles and cyclists which travel in the same direction shall be avoided at least within the conditions determined by the following equation. 
The compliance with this equation is required only for road users cutting
in, and only if the inserting road users were visible at least 0.72 seconds before cut-in (equation \eqref{eq:ttc_intrusion})".
\end{itemize}

\begin{equation}\label{eq:ttc_intrusion}
       TTC_{cut-in} \ge \frac{(v_{r} - v_f)}{2a} + \frac{1}{2}\tau + \tau_{reaction}
\end{equation}

Where $TTC_{cut-in}$ is the Time to-collision at the moment of the cut-in of the vehicle or cyclist by more than 30 cm in the lane in front of the ADS-equipped vehicle. 
The relative velocity will be positive if the vehicle with ADS functions is driving towards a slower intruding vehicle. 
$\tau$ is the time in seconds [s] to reach a desired deceleration rate ($a$) in [$m/s^2$] and $\tau_{reaction}$ is the reaction Time in seconds [s] necessary to initiate a brake response. 
$a$ is the Deceleration in meters per square second [$m/s^2$].

\begin{figure}[H]
\centering
 \includegraphics[width=0.6\textwidth]{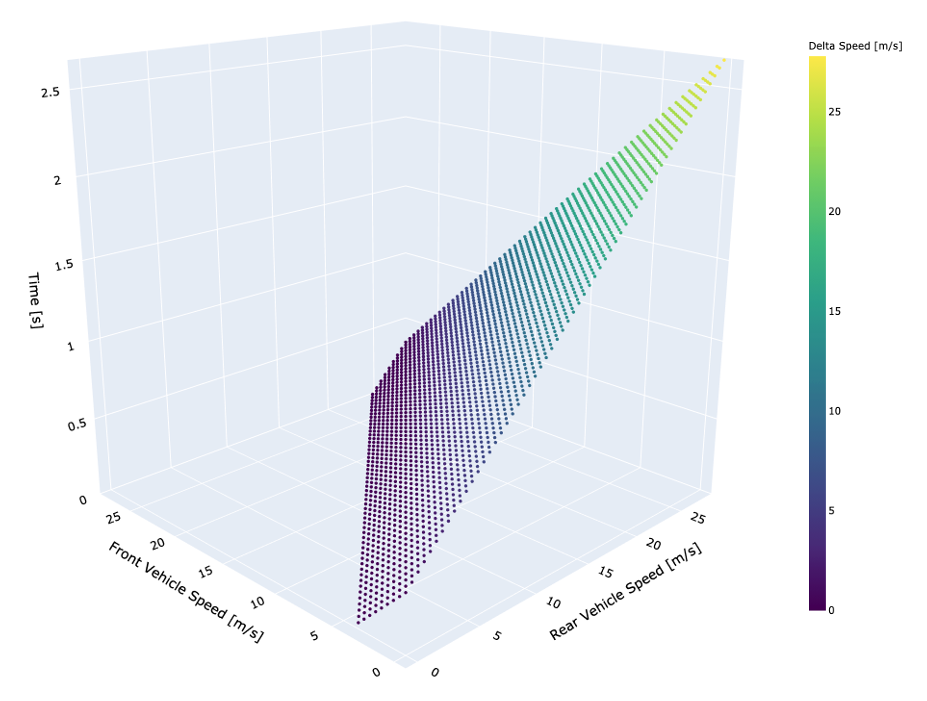}
\caption{TTC requirements for collision avoidance with cut-in vehicle}
\label{fig:ttc_cut_in}
\end{figure}

Figure \ref{fig:ttc_cut_in} shows the $TTC_{cut-in}$ requirements when a front vehicle with velocity $1.4\ m/s \le v_f \le 27.7\ m/s$ cuts in front of a rear vehicle (the ADS) that is traveling with $v_r \le 27.7\ m/s$ velocity. We can see that a maximum TTC value of 2.5s happens when the ADS is driving at 27.7 m/s (100 km/h) and the front vehicle performs a cut-in when driving at 1.4 m/s (5 km/h). 

While formulation \eqref{eq:ttc_intrusion} does account for the response time and braking capabilities of the ADS, which are important aspects for assessing its ability to avoid a collision, the arbitrary limitation that the performance requirement applies only when the cut-in happens 30 cm already into the ADS' lane means that common daily dangerous situations that don't conform to the arbitrary 30 cm requirement are thus not covered by the draft regulation.

In a cut-in scenario, it is critically important to consider the lateral component of the conflict as depending on the lateral velocity of the other road user performing the cut-in and the current longitudinal position and distance w.r.t. the ADS. 
In this way, the ADS could identify a cut-in manoeuvre much earlier and avoid all cut-in conflicts, not just those that conform to a lateral distance of 30 cm into the ADS' lane. 

Following the concepts introduced in \cite{shalev2017formal}, a dangerous situation is when both, the longitudinal and the lateral safety margins are violated. 
Which margin was violated last (longitudinal or lateral) dictates what the ADS should do in order to avoid a collision. 
For example, if the longitudinal safety margin was violated last, the ADS should apply a longitudinal brake to avoid a collision. 
If instead the lateral safety margin was violated last, this means that the cut-in happened at a very close longitudinal distance and therefore the ADS may not avoid a collision by braking longitudinally; it needs to perform a lateral manoeuvre. 
Such common driving situations does not appear to be addressed in the draft regulation.

\section{Discussion and Next Steps}
Any regulation supporting the deployment of ADS-equipped vehicles must not only promote the safety of the technology, but also ensure practicability and societal acceptance.  Regulatory performance requirements that would lead to vehicles being unable to support common daily traffic scenarios or drive so conservatively as to anger road users would fail on both criteria.

In this paper we have shown that the current draft regulation's reliance on arbitrary fixed TTC values will lead both to overly conservative behavior (particularly at intersections) and in many other cases, will result in crashes if the leading vehicle were to suddenly brake, or if road conditions were not ideal (e.g., wet, snow). 

There is also a substantial benefit to avoid ambiguity and possible confusion by further defining "sufficient space" and "sufficient time" terms.  What may be assumed to be "sufficient" by one ADS-equipped vehicle may be deemed "insufficient" by another leading to conflicts between automation systems for making different assumptions and calculations on such foundational performance characteristics.

Further, the lack of a specific definition on what it means for an ADS-equipped vehicle to "be cautious" and "avoid harsh braking" leaves large subjective loopholes that could be interpreted in any number of ways.  Regulation should provide clarity to industry.

Fortunately, the \RSS model provides an open and transparent technology-neutral way to explicitly define -- in a dynamic manner -- what constitutes "sufficient space" (and/or time), and precisely what it means for an ADS-equipped vehicle to be "cautious" and "avoid harsh braking."

While it may be tempting to define performance requirements that imagine an ADS-equipped vehicle could have no impact on other drivers, that logic is flawed, just as adding another human driver to an existing scenario will inevitably have an impact on the behavior of the other road users.
The art of driving is a multi-agent challenge with shared responsibilities, one where road users must negotiate with each other and make reasonably foreseeable assumptions about the reasonable worst-case behavior of the other agents.
Only through this artful dance does our transportation network work, and only if ADS-equipped vehicles are given the same consideration will we have both safe and practicable automated vehicles that will be embraced by the public.

\section{Conclusion}
In this paper, we analyzed the draft European Commission regulation for uniform procedures and technical specifications for the type-approval of motor vehicles with an automated driving system.  In the draft regulation, performance requirements on the DDT of an ADS-equipped vehicle are specified, covering traditional car following scenarios, lane change procedures, turning and crossings and emergency conditions.

Our analysis showed that the draft regulation as written, would result in ADS-equipped vehicles contributing negatively to the flow and safety of traffic, while leaving common every day traffic scenarios uncovered. Proposals addressing these issues were presented and discussed. 



\section{Contact details}
\begin{multicols}{2}

\begin{wrapfigure}{l}{21mm} 
\includegraphics[width=1in,height=1.2in,clip,keepaspectratio]{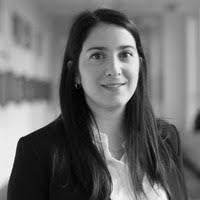}
\end{wrapfigure}

\textbf{Maria Soledad Elli}, joined Intel in 2017 as Data Scientist and since then she has been involved in several projects focused on automated vehicle's safety. She obtained her master’s degree in Data Science from Indiana University Bloomington in 2017, with focus on machine learning and computer vision applications. In 2013, she received her bachelor’s degree in Computer Engineering at the National University of Tucuman, Argentina. She worked until 2015 as a Software Engineer at the Aerospace and Government Division at INVAP SE, one of the leading Latin American corporations in applied high-tech. \\
\textbf{Email:} maria.elli@intel.com
\vfill\null
\columnbreak

\begin{wrapfigure}{l}{20mm} 
\includegraphics[width=1in,height=1.2in,clip,keepaspectratio]{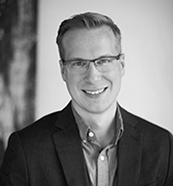}
\end{wrapfigure}

\textbf{Jack Weast}, is an Intel Fellow and a Vice President for Automated Vehicle Standards at Mobileye. In this role, Jack leads a global team working on AV safety technology and the related standards that will be needed to understand what it means for an AV to drive safely. Jack is the co-author of “UPnP: Design By Example”, and is the holder of over 40 issued patents with dozens pending.  Jack is an Adjunct Professor at Portland State University where he was recently inducted into the Portland State Maseeh College Academy of Distinguished Alumni in recognition of Jack’s achievements, leadership and service to the Engineering and Computer Science Profession, as well as to Society.\\
\textbf{Email:} jack.weast@intel.com

\end{multicols}
\clearpage
\bibliographystyle{unsrt}
\bibliography{references}
\end{document}